\documentclass[runningheads]{llncs}
\usepackage{times}
\usepackage{latexsym}
\usepackage{url}

\usepackage{booktabs} % For formal tables
\usepackage{multirow}
\usepackage{array}
\newcolumntype{C}[1]{>{\centering\let\newline\\\arraybackslash\hspace{0pt}}m{#1}}

\usepackage{amsmath}
\usepackage{float}
\restylefloat{table}
\usepackage{xcolor}
\usepackage{amssymb}
\usepackage{graphicx}
\usepackage[table]{colortbl}

\title{E2EET: From Pipeline to End-to-end Entity Typing \\ 
via Transformer-Based Embeddings}

\titlerunning{E2EET: From Pipeline to End-to-end Entity Typing}

\author{Michael Stewart \and Wei Liu}

\institute{The University of Western Australia \\
 \email{michael.stewart@research.uwa.edu.au, wei.liu@uwa.edu.au}}

\begin{document}
\maketitle
\begin{abstract}
  Entity Typing (ET) is the process of identifying the semantic types of every entity within a corpus. In contrast to Named Entity Recognition, where each token in a sentence is labelled with zero or one class label, ET involves labelling each entity mention with one or more class labels. Existing entity typing models, which operate at the mention level, are limited by two key factors: they do not make use of recently-proposed context-dependent embeddings, and are trained on fixed context windows. They are therefore sensitive to window size selection and are unable to incorporate the context of the entire document. In light of these drawbacks we propose to incorporate context using transformer-based embeddings for a mention-level model, and an end-to-end model using a Bi-GRU to remove the dependency on window size. An extensive ablative study demonstrates the effectiveness of contextualised embeddings for mention-level models and the competitiveness of our end-to-end model for entity typing.% introduce two models: a \textit{mention-level} model, featuring two novel attention mechanisms; and an \textit{end-to-end} model.  We show that our mention-level model outperforms existing state-of-the-art techniques while our end-to-end model is capable of outperforming mention-level models.
\end{abstract}

\section{Introduction}

Entity Typing (ET) is the process of identifying the semantic types of every entity within a corpus. In contrast to Named Entity Recognition, where each token in a sentence is labelled with zero or one class label, ET involves labelling each entity mention with one or more class labels, which are typically arranged in a hierarchy~\cite{ling2012fine}. These fine-grained class labels encapsulate more semantic information than singular labels, and allow for entities to be labelled with mutually exclusive types. The results obtained by ET are therefore highly valuable for many downstream natural language processing tasks such as  information extraction~\cite{sarawagi2008information}, knowledge graph construction~\cite{stewart2019icdm}, and text mining~\cite{stewart2017interactive}.

% Modern ET systems are trained to predict the set of labels associated with each entity mention in a corpus. 

Despite the widespread success of entity recognition, research into effective entity typing is still ongoing. End-to-end entity typing, whereby every token is labelled with zero or more type(s), is considerably more challenging than entity recognition as it is a multi-class, multi-label task~\cite{tsoumakas2009mining}. This difficulty has resulted in the majority of state-of-the-art entity typing systems assuming the segmentation step and operating only at the mention-level. In other words, each entity mention in the dataset \textit{has already been identified} and is labelled with one or more semantic classes. %, in contrast to Entity Recognition which is multi-class, single-label. 

% Modern ET models aim to predict the set of types associated with each entity mention. The input data to the model must be already segmented, such that the entity mentions are labelled and each entity always has at least one type. 

%mention-level typing research primarily focuses on the typing of large, widely-available corpora such as Wikipedia and news reports. MT models are typically designed to perform well on these datasets, but they are seldom built with other less popular domains in mind. %The fact that existing MT systems rely upon the input data being already segmented prevents these models from operating on non-segmented data, such as industry-specific datasets or social media data. An end-to-end model that performs both the segmentation stage and the classification stage would highly beneficial to a wide variety of downstream NLP applications, such as text mining from industrial data.

% Keep?
% There have recently been great advances in the area of language representation. State-of-the-art embedding models are typically context-dependent, meaning the embeddings of each word are generated with respect to the word's surrounding context. ELMo \cite{peters2018elmo} is based upon a Bidirectional Long Short-Term Memory (LSTM) architecture. BERT~\cite{devlin2018bert}, on the other hand, adopts an entirely feed-forward approach using a bidirectional transformer and has improved the state-of-the-art in a wide range of NLP tasks.

There are a number of limitations that arise from performing entity typing at the mention level, as opposed to end-to-end, however. Firstly, the state-of-the-art technique for mention-level entity typing is to train on fixed windows of tokens centered around the entity mention using a three-part Bi-LSTM~\cite{dong2015hybrid}, in an attempt to prevent the model from training on irrelevant information. This not only results in the model being highly sensitive to the size of the context window, but also means that it will never be able to incorporate the contextual information outside the window when it is actually important. An end-to-end model employing a bidirectional gated recurrent unit (GRU) would be capable of learning to harness this context effectively via the forget gate~\cite{cho2014learning}. Secondly, to perform entity typing from scratch with a mention-level system one must also train a segmentation model and combine both systems, which is slower and a more complicated pipeline than training one end-to-end system. We show that an end-to-end model is capable of  outperforming a mention-level model given the right architecture, whilst also simplifying the task. 

%Several limitations hinder the wider adoption of existing mention-level Entity Typing systems in real-world applications. 

Existing models are also hindered by their input representations. The input representation of each word in state-of-the-art mention-level typing models is generated using context-independent embedding models, such as GloVe~\cite{pennington2014glove}. The effectiveness of state-of-the-art context-dependent embedding models, such as BERT~\cite{devlin2018bert}, in entity typing has not yet been investigated despite its proven success in many other NLP tasks such as Named Entity Recognition. In addition to explicitly learning a context representation using an end-to-end model, we also investigate the effectiveness of contextualised word embeddings on mention-level entity typing.

% In other words, entity typing is treated as a downstream task of entity recognition. This severely limits the applicability of Entity Typing models in domains that cannot be easily segmented into entities and non-entities. Short text domains such as industrial data and social media, for example, are rife with non-standard tokens such as spelling errors and domain-specific terms and as such are extremely challenging to segment using pre-trained entity recognition models. An end-to-end model that performs both the segmentation stage and classification without relying on handcrafted features would be highly beneficial to a wide variety of real-world NLP applications, such as text mining from industrial data. It would also be more widely applicable to other languages such as Chinese and Japanese where token segmentation is non-trivial.

We therefore carry out an extensive ablative study to demonstrate effectiveness of contextualised embeddings for mention-level entity typing, and show the competitiveness of end-to-end entity typing despite it being a more challenging learning task. We accomplish this by introducing two models: a \textit{mention-level} model which embeds the left, right, and mention contexts using BERT, and an  \textit{end-to-end} entity typing (E2EET) model that determines the type(s) of all tokens in a sentence. 

In this paper we describe our two models in detail in Section~\ref{sec:et-model}. In Section~\ref{sec:results} we evaluate our mention-level model and show that it outperforms state-of-the-art mention-level entity typing models. We also show that E2EET is effective on clean datasets and is capable of outperforming  mention-level entity typing models despite not knowing which tokens in each sentence are entities apriori.

% , as opposed to state-of-the-art embedding models such as BERT~\cite{devlin2018bert}.

%\filler{Why is this challenging?}

%Entity recognition models are typically single-class, multi-label. In entity recognition, mentions are labelled with exactly one class, while non-entities are labelled with a special ``outside'' class. Entity typing systems, on the other hand, label already-segmented entity mentions with their corresponding class(es), typically making use of the forward and backwards contexts of each mention. Some systems treat entity typing as a single-class multi-label task, labelling mentions with exactly one type.

%The concept of treating entity typing as a multi-class, multi-label problem was first introduced by Ling \& Weld~\cite{ling2012fine}.

%Recent entity typing models do not typically perform the segmentation step, and are instead trained on data that has already been segmented via distant supervision. The goal of modern entity typing models is to predict the set of labels associated with each entity mention, given its left and right context.

%\filler{Talk about BERT and how it has not been used either?}

%To the best of our knowledge, no system is currently capable of performing E2E ET without relying on handcrafted features to perform the segmentation step. 

%Despite the prevalence of ET in the literature, very few systems perform true end-to-end entity typing. FIGER~\cite{ling2012fine} is unique in that it performs both segmentation and classification, but relies heavily on handcrafted features. 

\section{Related Work}

\label{sec:relatedwork}
%Entity recognition is historically a single-class, multi-label classification problem. 

% Fine-grained entity typing

Initial entity typing research treated the task as a multi-class, multi-label classification problem, i.e. identify the type(s) of every entity within a document. The Fine-Grained Entity Recognition (FIGER)~\cite{ling2012fine} model follows a pipeline-based approach: it first identifies the entity mentions via a segmentation step, and then predicts a list of class labels (types) associated with each mention. The segmentation is performed using a conditional random field (CRF) trained on a variety of handcrafted features such as token length and contextual bi-grams. Label prediction is performed using a multi-layer perceptron. FIGER's reliance on handcrafted features to perform segmentation makes it unfeasible for domains and applications where these features are not readily available.

% 2. Three-part BiLSTM

More recent research focuses on mention-level entity typing. In contrast to the two-staged pipeline of FIGER, which performs entity segmentation (i.e. entity recognition) followed by label prediction, mention-level models are trained on already-segmented data~\cite{murty2018hierarchical} and aim to predict the type(s) of each entity mention given its context. 

State-of-the-art entity typing models typically employ a three-part Bi-directional Long Short-Term Memory Model (Bi-LSTM). The left, right, and mention contexts are each fed through a Bi-LSTM layer to obtain an encoded representation, which is then decoded and fed through a linear layer to obtain a set of labels for the corresponding mention. This architecture was introduced as part of the Hybrid Neural Model (HNM)~\cite{dong2015hybrid}, which comprises two components: a recurrent-based mention model that obtains a vector representation of the entity mention given its context, and a context model that generates a single vector for both the left and right contexts of the mention. The output from the mention model and context model is concatenated and fed through a softmax layer to obtain a probability distribution over all possible types. The type with the highest probability is selected as the prediction.

Multi-Instance Entity Typing from Corpus (METIC)~\cite{xu2018metic} also employs a three-part Bi-LSTM. Once the input has been embedded via GloVe~\cite{pennington2014glove} and passed through the Bi-LSTM layers, the output is concatenated and two constraints form the basis of an integer linear programming model which is applied to the output of the final dense layer. The \textit{type disjointness constraint} ensures that an entity is not labelled as two mutually exclusive types. The mutual exclusivity of each type are determined via an external knowledge base. The \textit{type hierarchy constraint} ensures that an entity is not labelled as a certain type if it is not also labelled as that type's parent category.

In contrast to other mention-level entity typing systems, Automatic Fine-Grained Entity Typing (AFET) ~\cite{ren2016afet} does not employ a Bi-LSTM. It instead introduces a novel, heuristic-based method to separate clean and noisy mentions, whilst also introducing hierarchy-based partial label embeddings to improve performance. AFET takes advantage of the noisy labels in the dataset; mentions are separated into clean and noisy sets depending on whether their ground truth labels form a single path in the category hierarchy, i.e. are not mutually exclusive. The loss function of the model differs for each training example depending on whether it is from the clean or noisy set. AFET notably relies on handcrafted features (such as POS tag and Brown Cluster), unlike other systems, which limits its functionality on datasets that have not been labelled by hand.

In summary, FIGER is a pipeline-based entity typing system that is heavily reliant on feature collection. HNM, METIC and AFET are all deep learning-based mention-level entity typing systems. HNM and METIC use a three-part Bi-LSTM for representation learning, while AFET relies on handcrafted features.

\section{Entity Typing Models}
\label{sec:et-model}
This paper introduces two Entity Typing models: a \textit{mention-level} model, which determines the type(s) of an entity given an entity mention and its surrounding context, and an \textit{end-to-end} model (E2EET), which determines the type(s) (if any) of all tokens in a sentence. We begin this section by providing an overview of the embedding layer that is common between both models, and then explain each model in detail. % TODO: describe overall architecture

The embedding layer plays a crucial role in our models, allowing for a context-dependent, deep representation of input tokens. To facilitate such a representation, we use BERT~\cite{devlin2018bert}. BERT is based upon the encoder stack of the bidirectional transformer model~\cite{vaswani2017attention}, an encoder-decoder model structure that combines feed-forward layers with a multi-headed attention mechanism. In contrast to other state-of-the-art context-dependent embedding models, such as ELMo~\cite{peters2018elmo}, BERT learns deep bidirectional representations by performing a procedure known as the ``masked language model''. For every input sentence to the model some number of terms are masked at random, and the model learns to predict the original terms.

BERT embeddings allow our models to incorporate valuable contextual information in the embedding layer, as opposed to being learned via a recurrent layer as is common in existing entity typing systems. The embeddings generated by BERT and fed into our model are context-dependent, meaning the embeddings of each word are generated with respect to its surrounding context. Polysemous words are embedded according to their canonical meaning, providing a richer representation than the context-independent embedding models that are currently used in many state-of-the-art entity typing models. 

An important distinction between BERT and other embedding models is that it is trained at the ``wordpiece'' level (also known as Byte Pair Encoding)~\cite{sennrich2015neural}, as opposed to the word level. A single unknown token must first be tokenized into pieces prior to being fed through the BERT model. For example, ``Johanson'' becomes ``Johan'' ``\#\#son''. %BERT also requires special \texttt{[CLS]} and \texttt{[SEP]} tokens at the start and end of the sentence respectively.

\subsection{Mention-level model}

\begin{figure}[!ht]
    \begin{center}
        \includegraphics[width=0.6\columnwidth]{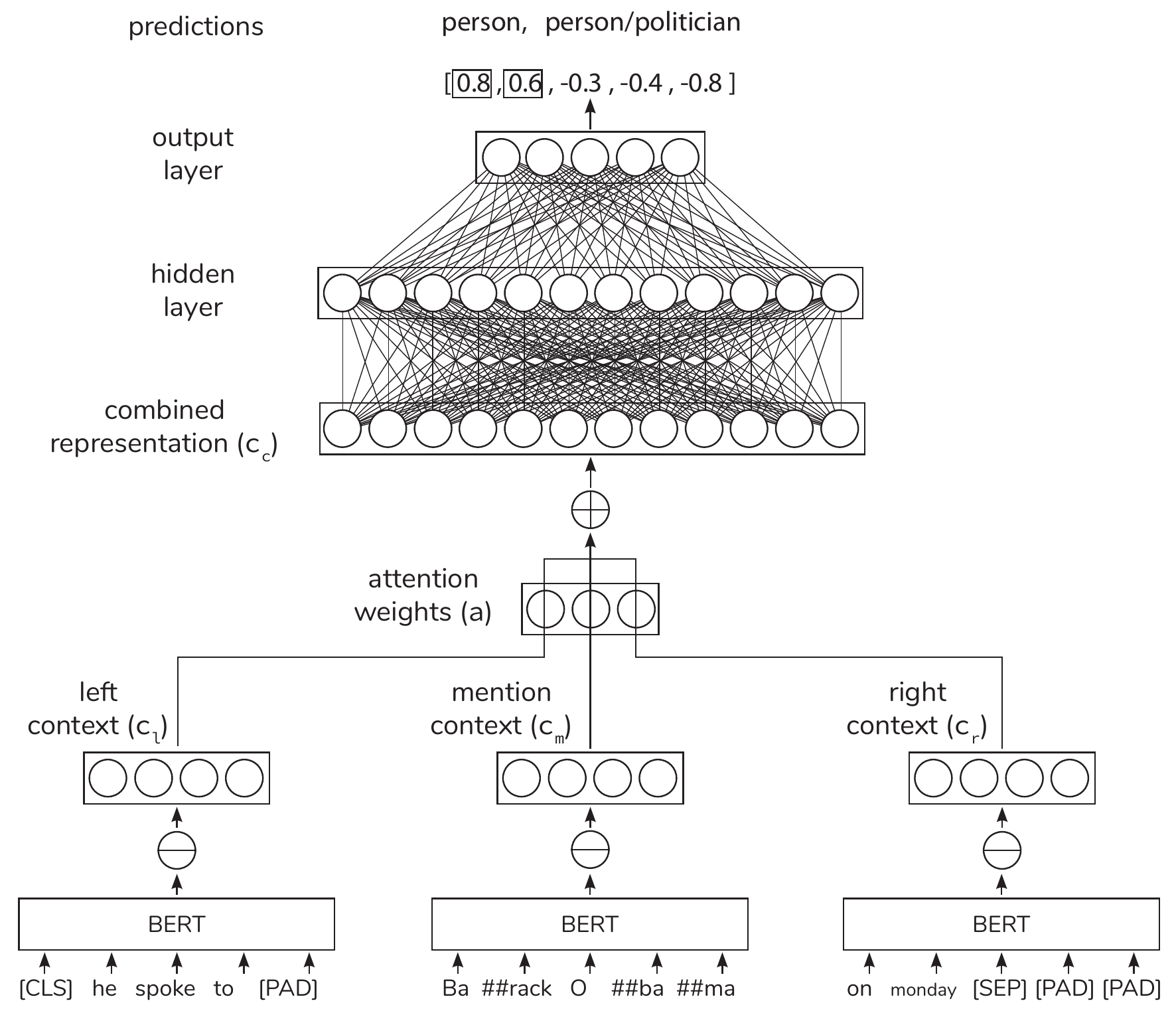}
        \caption{The architecture of the mention-level model. Circles with horizontal lines through them denote that the weights are averaged, while the crossed circle indicates concatenation.}
        \label{fig:ml-diagram}
    \end{center}
\end{figure}

The mention-level model, as shown in Figure~\ref{fig:ml-diagram}, predicts the label(s) of a given entity mention and its surrounding context. It accomplishes this using a three-part context model, consisting of the left, right, and mention context, inspired by the state-of-the-art mention-level systems discussed in Section~\ref{sec:relatedwork}. However, our model does not employ a recurrent neural network, and instead uses two feed forward layers: one to learn an encoded representation of the combined left, right, and mention contexts, and another to map this encoded representation back to the label space.

\subsubsection{Context vectors}

We first build the left and right context windows by taking $W$ wordpieces to the left and right of the mention, respectively, where $W$ is a fixed context window size. For the mention context window, we take the first $W$ wordpieces of the mention. If any vector is not of length $W$, it is padded with  \texttt{[PAD]} tokens. If its length is greater than $W$, excess wordpieces are trimmed. Each of the left, right, and mention context windows are then encoded via a pre-trained BERT model to obtain three embedding matrices of size $W \times d$, where $d$ is the embedding dimension. We then take the average across each of these matrices, yielding three vectors of size $d$. These three vectors are hereby denoted as the left, right, and mention context vectors $c_l$, $c_r$, and $c_m$.

%In the example in Figure~\ref{fig:ml-diagram}, the context window size is 5, and the embedding dimension is 4. Each context window is embedded via BERT, and the embeddings are averaged to form the left, right, and mention contexts.

\subsubsection{Attention mechanisms}
\label{subsec:attention}
The mention-level model may be augmented with one of two attention mechanisms: \textit{scalar} and \textit{dynamic}.

The \textit{scalar} attention mechanism learns the extent to which each context (left, right, and mention) is important when predicting the labels of each mention. The weights of each context are then multiplied according to their relevance to the task. To do this, a scalar value $\widetilde{a_i}$ is learned for each context vector $c_l$, $c_r$, and $c_m$. We normalise the attention weights using the softmax operation so that they sum to 1 (here, $C$ represents the contexts $\{l, r, m\}$). The weights of each layer are multiplied by the corresponding attention value and are then concatenated to form $c_c$ as shown in Figure~\ref{fig:ml-diagram}.

% \begin{subequations}
% \begin{gather}
%     \widetilde{a_i}=\frac {e^{a_{i}}}{\sum _{i=1}^{C}e^{a_{i}}} \\
%     c_l = c_l \cdot \widetilde{a_1} \\
%     c_r = c_r \cdot \widetilde{a_2} \\
%     c_m = c_m \cdot \widetilde{a_3} \\
%     c_c = c_l \odot c_r \odot c_m
% \end{gather}
% \label{eqn:attention}
% \end{subequations}

The three attention weights are applied to every mention regardless of the mention itself. This results in low complexity but has the downside of assuming that every mention will benefit from the same attention weights. For example, if the attention value is high for the left context and low for the mention and right contexts, and a mention has no left context (i.e. it is at the start of the sentence), the predictions for this particular mention may be adversely affected by the attention mechanism.

% \begin{figure}[!ht]
%     \begin{center}
%         \includegraphics[width=0.6\columnwidth]{diagrams/ml-diagram-dynamic}
%         \caption{The relevant components of the mention-level model that are changed when using the dynamic attention mechanism.}
%         \label{fig:ml-diagram-dynamic}
%     \end{center}
% \end{figure}

In light of this issue, we propose a \textit{dynamic attention} mechanism. Rather than learning one weight per context layer, the dynamic variant uses a much simpler feed-forward network to assign weights to each context layer based upon the mention context. The inputs to this layer are the averaged embeddings across each wordpiece in the mention context. The output is a vector of three weights corresponding to the left, right and mention context, which are softmaxed and applied to each layer $h_l$, $h_r$ and $h_m$.% as per equation~\ref{eqn:attention}.

The dynamic attention mechanism allows for the predictions of polysemous mentions to be more heavily influenced by their surrounding context, whilst also reducing irrelevant contextual information for non-polysemous mentions. For example, given the mention ``Apple'' (which could be a company or fruit depending on the context), the normalised weights of the attention layer's output might be $[0.45, 0.45, 0.1]$ for the left, right, and mention contexts respectively. Given another example, where the entity types may be easily inferred from the mention itself (such as ``Barrack Obama''), the normalised weights might be $[0.1, 0.8, 0.1]$.

%It should also be noted that it is possible to run the model without an attention mechanism, in which case the attention weights are effectively $[0.33, 0.33,  0.33]$.

\subsubsection{Hidden and output layers}

After being multiplied by the attention weights, the three context vectors are concatenated to form the combined representation $c_c$. This vector is fed through a linear layer, followed by a ReLU activation function. The outputs of this layer are fed through one final layer to obtain the output vector, $x$, which  contains one weight corresponding to each label $\in N$, where $N$ is the set of all labels.

% Given $h$ is the output of the ReLU function, $A$ is a weight matrix that is learned, and $b$ is the bias:

% \begin{equation}
% x = hA^{\top} + b
% \label{eqn:linear-2}
% \end{equation}

%The output vector, $x$, contains one weight corresponding to each label $\in N$, where $N$ is the set of all labels.

\subsubsection{Loss function}

After performing the sigmoid function to normalise the weights to between 0 and 1, the loss of the model is calculated using binary cross entropy :

\begin{subequations}
\begin{gather}
% y\prime = S(x) =  \frac{1}{1 + e^{-x}} \\
l_n = -y_{n} log(y\prime_{n}) - (1 - y_{n}) log(1 - y\prime_{n}) \\
loss = \frac{\sum_{n \in N}{l_n}}{|N|}
\end{gather}
\label{eqn:loss-2}
\end{subequations}

Here, $y_n \in\{0, 1\}$ is the correct label of the class of index $n$, $\{y\prime_n \in \mathbb{R}\, |\, 0 \leq y\prime_n \leq 1\}$ is the prediction score associated with the class of index $n$, and $N$ is the set of labels.

\subsubsection{Prediction layer}

Given a set of prediction weights $y\prime$ across each label $n$, the prediction layer outputs $1$ when $y\prime_n > 0.5$ and $0$ when $y\prime_n <= 0.5$. 

One minor difference between the mention-level model and the end-to-end model is that in mention-level typing, each entity mention is guaranteed to have at least one label. To address this issue we adjust the prediction layer of our mention-level model to output its highest-weighted label in the event that no prediction weights are $> 0.5$. 

\subsection{End-to-end model (E2EET)}

\begin{figure*}[!ht]
    \begin{center}
        \includegraphics[width=0.75\linewidth]{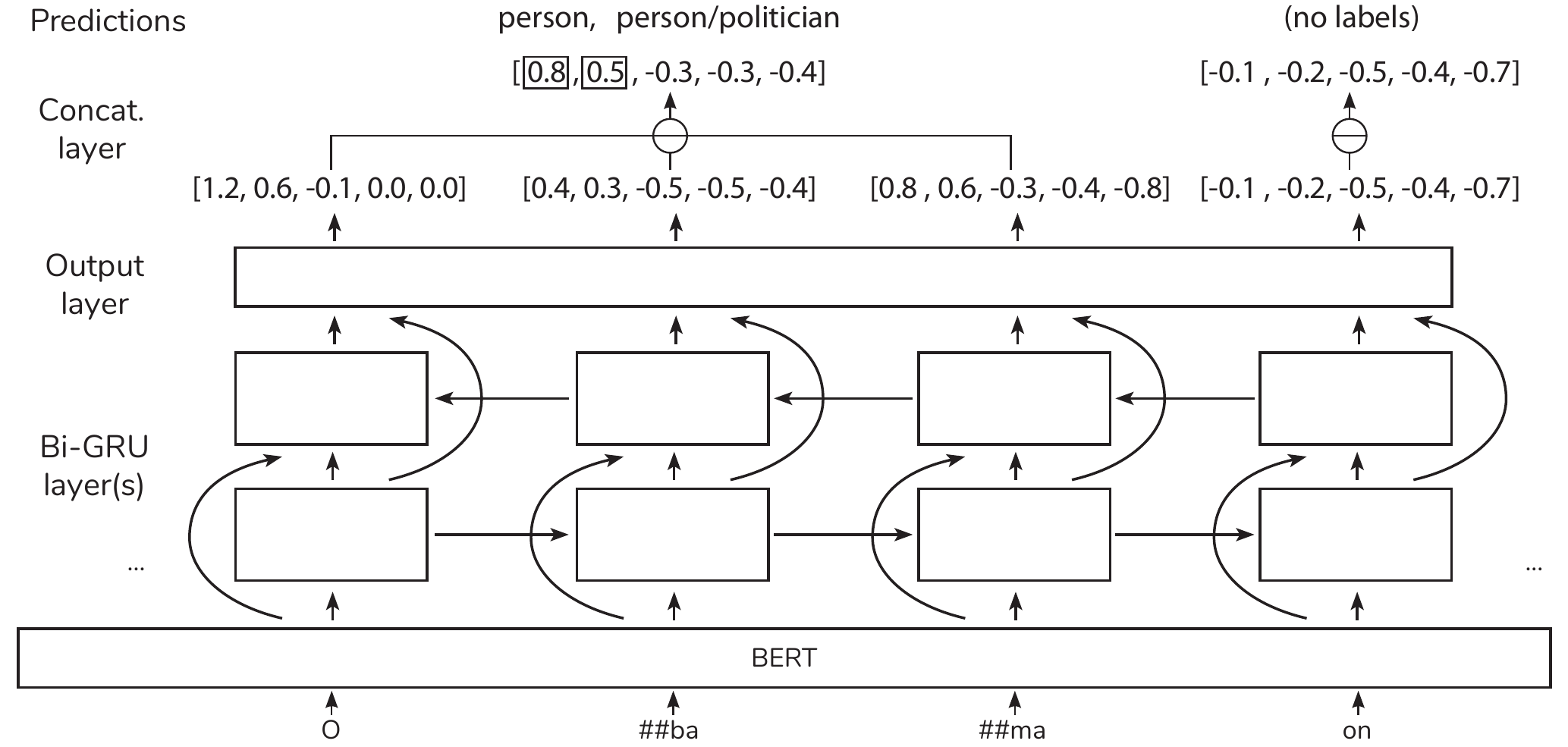}
        \caption{The architecture of the end-to-end model (E2EET). The remaining wordpieces at either side of the four example wordpieces are not included in the diagram for brevity.}
        \label{fig:e2e-diagram}
    \end{center}
\end{figure*}

The end-to-end entity typing model (E2EET), as shown in Figure~\ref{fig:e2e-diagram}, predicts the type(s) of every token in a sentence. In contrast to the mention-level model, it does not require the entities to be segmented and can operate on a dataset composed purely of raw text. Rather than taking an entity mention and its context as input, it takes an entire sentence as input and outputs a set of zero or more labels for each token in the sentence. % The context vectors and hidden layers are identical to the mention-level model, but no attention mechanism is used due to the lack of left, right and mention contexts.

%While end-to-end entity typing is significantly more challenging to perform than mention-level entity typing, it does not require the entities to be segmented in the evaluation dataset. This means that the model can perform entity typing on a dataset composed purely of raw text, with no need to perform entity segmentation beforehand.

E2EET is similar in architecture to our mention-level model, but uses a bidirectional gated recurrent unit (GRU)~\cite{cho2014learning} instead of a feed-forward network. This allows for both forward and backwards contexts (independent of window size) to be taken into account when predicting the label(s) of each token.

\subsubsection{Loss function}

The loss of E2EET is calculated using binary cross entropy in the same manner as the mention-level model as per Equation~\ref{eqn:loss-2}. However, rather than averaging across a single set of label predictions, the loss of E2EET is averaged across the predictions of all wordpiece tokens $T$ in the current batch.

% \begin{subequations}
% \begin{gather}
% loss = \frac{\sum_{t=1}^{T}{loss(t)}}{|T|}
% \end{gather}
% \label{eqn:loss-e2e}
% \end{subequations}

\subsubsection{Concatenation layer}

As opposed to many embedding models, BERT operates at the wordpiece level. The outputs of the E2EET must be a set of labels per token, not per wordpiece. Our concatenation layer therefore takes the average predictions of each wordpiece label corresponding to a particular word as shown in Figure~\ref{fig:e2e-diagram}.

% During the data loading stage, we construct a map between token indices and their corresponding wordpiece indexes in each sentence as described by~\cite{devlin2018bert} in the BERT documentation\footnote{https://github.com/google-research/bert}. For example, given that the sequence of tokens \texttt{["John", "Johanson", "'s", "house"]} is tokenized to \texttt{["[CLS]", "john", "johan", "\#\#son", "'", "s", "house", "[SEP]"]}, the map, hereby denoted as $m$, for this sentence is constructed as \texttt{[1, 2, 4, 6]}. 

% The concatenation layer works as follows. Given a set of predictions $y\prime$ for each wordpiece in a sentence, the layer takes the average of every prediction that corresponds to the same token according to $m$. In the above example, the four token predictions are based on the predictions corresponding to the first wordpiece (\texttt{"john"}), the average of the second and third wordpiece (\texttt{"johan"} and \texttt{"\#\#son"}), the average of the fourth and fifth wordpiece (\texttt{"'"} and \texttt{"s"}), and the sixth wordpiece (\texttt{"house"}) respectively. This allows for the model to output token-level predictions despite being trained at the wordpiece level.

%\filler{Hierarchical encoding}

%\filler{RELU, Dropout layers}

\section{Experiments}

\subsection{Datasets}

\newcolumntype{x}[1]{>{\centering\arraybackslash\hspace{0pt}}p{#1}}

\begin{table}[H]
\centering
\begin{tabular}{l | x{40pt} | x{40pt} | x{40pt}| x{40pt} | x{40pt} | x{40pt}}

 & \multicolumn{3}{c|}{\textbf{Original}} &  \multicolumn{3}{c}{\textbf{Modified}} \\ \hline
\textbf{Name} & \textbf{\# Train}  & \textbf{\# Dev} & \textbf{\# Test} & \textbf{\# Train}  & \textbf{\# Dev} & \textbf{\# Test}  \\ \hline
%\multicolumn{5}{c}{Word-level model} \\

Wiki        & 1,505,241 & 434   & 434 & 50,000 & 434 & 434 \\
Ontonotes   & 79,456    & 8,828 & 1,312 & 1,048 & 132 & 132 \\ 
BBN       & 29,466    & 3,273 & 6,431 & 5,143 & 644 & 644 \\ 

\end{tabular}
\caption{\label{tab:datasets-orig} The number of sentences in the original datasets (used to evaluate our mention-level model) and modified datasets (used to evaluate the end-to-end model).
  }
\end{table}

We evaluate our mention-level model on the benchmark datasets provided by~\cite{ren2016afet}: Wiki, Ontonotes, and BBN\footnote{https://github.com/INK-USC/AFET}. We use a portion of the training datasets as validation sets (434 for Wiki, and 10\% for Ontonotes and BBN). These datasets are summarised in Table~\ref{tab:datasets-orig} in the \textit{Original} column.

%Initial data exploration found that the training sets of the Ontonotes and BBN datasets are of little value to an end-to-end entity typing system because the labels assigned to each token are almost universally incorrect. This is supported by~\cite{xu2018metic}, who found that the provided datasets are not the same datasets as in the original paper. 

Initial data exploration of the training sets of the Ontonotes and BBN datasets found that the data contained a high proportion of incorrect labels. The testing sets, however, appear to be free from error as a result of being manually annotated~\cite{shimaoka2016neural}. We therefore created our own versions of these datasets, hereby known as the ``modified'' datasets and prefixed with an ``M''. The datasets are summarised in Table~\ref{tab:datasets-orig} in the \textit{Modified} column, 

The training set of the Wiki dataset is relatively clean when compared with the training sets of the original Ontonotes and BBN datasets. However, we found that due to the complexity of our end-to-end model, it was necessary to trim the large Wiki dataset in order for it to fit in memory. We therefore constructed the M-Wiki dataset by taking the first 50,000 documents of the original ~1.5 million document training set as the new training set and the following 434 as the new validation set. The test set is the same as in the original dataset. For the M-Ontonotes and M-BBN datasets, the training sets comprise the first 80\% of the test data. The validation sets comprise the following 10\%, and the test sets comprise the remaining 10\%.

%Table~\ref{tab:datasets} shows the number of documents in each of the modified datasets.

\subsection{Model parameters}

After parameter tuning we found that the best performance on the development set for both of our models were achieved with a learning rate of 0.0001, a hidden dimension size of 768, and 0.5 dropout prior to the final layer. The models were optimised using ADAM. The batch size was 100 for the mention-level model and 10 for the end-to-end model. For the mention-level model we used a context window size of 10 for the left, right, and mention contexts. The end-to-end model was trained with a max sequence length of 100, allowing for 99\% of the data to be included without dramatically increasing training time.

% \subsubsection{End-to-end model}

% The end-to-end model was trained using the same optimisation technique, learning rate, dropout, and hidden dimension size to the mention-level model. The batch size was set to 10. The model was trained with a max sequence length of 100, allowing for more than 99\% of the data to be included without dramatically increasing training time. We removed all documents in each dataset that contained more than 100 tokens after wordpiece tokenisation.

\subsection{Embedding techniques}

In order to evaluate the effectiveness of the BERT embeddings in our models, we evaluate our end-to-end model with four different embedding techniques. \textit{Uniform}, the baseline, assigns a uniform distribution of embedding weights for each token. The GloVe~\cite{pennington2014glove} embeddings are pretrained on Wikipedia 2014 + Gigaword 5\footnote{https://nlp.stanford.edu/projects/glove/}. The Word2Vec embeddings are pretrained on the Wikipedia corpus\footnote{https://wikipedia2vec.github.io/wikipedia2vec/pretrained/}. The embedding dimension of each of these techniques was 300. The BERT embeddings are generated from the pre-trained \texttt{BERT\textsubscript{BASE}, Cased} model\footnote{https://github.com/google-research/bert}, which provides embeddings of dimension 768. We used Bert-as-service\footnote{https://github.com/hanxiao/bert-as-service} to embed the sentences per-batch. We did not fine-tune BERT on our datasets as we found it did not provide a performance improvement.

\subsection{Evaluation metrics}

We evaluate our model using three standard metrics for entity typing systems in terms of F1 scores: Strict Accuracy, Loose Macro, and Loose Micro score, described in detail by~\cite{ling2012fine}. \textit{Strict Accuracy} only considers the prediction of a token correct when the set of predicted classes matches the set of ground truth classes exactly.  \textit{Loose Macro} calculates the scores for matching subsets at the entity level, individually for each entity mention, whereas \textit{Loose Micro} computes the score at the corpus level, and the score is averaged across all entities. Loose Macro tends to be penalised when new unseen categories appear in the test set, and is therefore sensitive in unbalanced datasets. In such cases, Loose Micro is a fairer metric.

\subsection{Baseline systems}

We compare our mention-level model to the state-of-the-art systems evaluated in~\cite{xu2018metic}:

\begin{itemize}
    \item AFET~\cite{ren2016afet}: A neural mention typing model that uses handcrafted features.
    \item HNM~\cite{dong2015hybrid}: A hybrid neural model for mention typing.
    \item HNM-ML: As above but adapted for multi-label typing.
    \item METIC~\cite{xu2018metic}: A neural model that uses a combination of Bi-LSTMs and integer linear programming.
\end{itemize}

%We do not compare our system to Metic-MT-Filtered from \cite{xu2018metic} because it was trained on datasets that had been cleaned.h

\section{Results}
\label{sec:results}

% Our results investigate the following:

% \begin{enumerate}
%     \item \textbf{mention-level typing performance}: How does our mention-level model compare to existing mention-level typing systems?
% 	\item \textbf{End-to-end performance}: How well does our end-to-end model perform when compared to the mention-level model?

% \end{enumerate}

\begin{table*}[!t]
\centering
\begin{tabular}{l | c | c | c | c | c | c | c | c | c}

\multirow{2}{*}{\textbf{Model}} & \multicolumn{3}{c|}{\textbf{Wiki}}  & \multicolumn{3}{c|}{\textbf{Ontonotes}}  & \multicolumn{3}{c}{\textbf{BBN}} \\ \cline{2-10}
  & Acc & Ma-F1 & Mi-F1 & Acc & Ma-F1 & Mi-F1 & Acc & Ma-F1 & Mi-F1 \\ \hline
%\multicolumn{5}{c}{Word-level model} \\

% Default                 & 0.7421 & 0.1651 & 0.7127 & 0.9591 & 0.2857 & 0.8193 & 0.9031 & 0.0570 & 0.4506 \\

\cellcolor{gray!10}{AFET}                    & \cellcolor{gray!10}{0.205} & \cellcolor{gray!10}{0.616} & \cellcolor{gray!10}{0.600} & \cellcolor{gray!10}{0.348} & \cellcolor{gray!10}{0.620} & \cellcolor{gray!10}{0.559} & \cellcolor{gray!10}{0.638} & \cellcolor{gray!10}{0.698} & \cellcolor{gray!10}{0.710} \\ \hline
% 
%\cellcolor{gray!10}{AFET}                    & \cellcolor{gray!10}{0.205} & \cellcolor{gray!10}{0.616} & \cellcolor{gray!10}{0.600} & %\cellcolor{gray!10}{\textbf{0.348}} & \cellcolor{gray!10}{\textbf{0.620}} & \cellcolor{gray!10}{\textbf{0.559}} & %\cellcolor{gray!10}{\textbf{0.638}} & \cellcolor{gray!10}{0.698} & \cellcolor{gray!10}{0.710} \\ \hline
HNM                     & 0.442 & 0.670 & 0.616 & \cellcolor{gray!10}{\textbf{0.344}} & 0.527 & 0.479 & 0.216 & 0.558 & 0.521  \\ \hline
HNM-ML                  & 0.471 & 0.657 & 0.683 & 0.267 & 0.503 & 0.493 & 0.469 & 0.682 & 0.680  \\ \hline
%MT-SP                   & 0.469 & 0.661 & 0.680 & 0.337 & 0.556 & 0.517 & 0.497 & 0.708 & 0.702  \\ \hline
METIC                   & 0.528 & 0.711 & \textbf{0.718} & 0.256 & 0.524 & 0.508 & 0.492 & 0.717 & 0.710 \\ \hline \hline
% METIC (F)               & 0.551 & 0.745 & 0.735 & 0.313 & 0.571 & 0.539 & 0.520 & 0.718 & 0.713  \\ \hline

ML (None)               & \textbf{0.528} & 0.712 & 0.688 & 0.286 & 0.562 & 0.523 & \cellcolor{gray!10}{\textbf{0.504}} & 0.734          & \textbf{0.727} \\ \hline
ML (Scalar)             & 0.519 & 0.715 & 0.685 & 0.278 & 0.555 & 0.523 & 0.498 & \textbf{0.735} & 0.724 \\ \hline
ML (Dynamic)             & 0.517 & \textbf{0.719} & 0.685 & 0.283 & \cellcolor{gray!10}{\textbf{0.577}} & \cellcolor{gray!10}{\textbf{0.537}} & 0.493 & 0.733          & 0.718 \\ \hline

%Ours (E2E)             & 0.316 & 0.059 & 0.516 & \textbf{0.610} & 0.170 & \textbf{0.749} & \textbf{0.702} & 0.296 & \textbf{0.844}  \\

\end{tabular}
\caption{\label{tab:performancecomparison} A comparison of our mention-level (ML) model, with three different attention mechanisms, against state-of-the-art mention-level typing systems. The system using handcrafted features, AFET, is highlighted in grey. Grey cells with bold text indicate the best performing model not using handcrafted features. Non-highlighted cells with bold text are the best performing model even when compared to systems using handcrafted features.
  }
\end{table*}

\subsection{Mention-level model performance}

\label{subsec:mentiontypingperformance}

Our first set of investigations is to determine how our mention-level model's performance compares to existing systems. We also investigate the effectiveness of the proposed attention mechanisms.

Table~\ref{tab:performancecomparison} shows the results of our model when compared to state-of-the-art mention-level models, with results for existing systems supplied by~\cite{xu2018metic}. AFET, which relies on handcrafted features, is highlighted in grey. 

Our model outperforms METIC, the top-performing system that does not rely on handcrafted features, in every experiment except the micro-F1 metric on the Wiki dataset. We attribute the success to the combination of the context-dependent BERT embedding vectors and the three-part context model. In contrast to existing state-of-the-art systems, our model is able to encapsulate contextual information via the context-dependent embeddings provided by BERT's transformer-based architecture.

Despite not relying on handcrafted features like AFET does, our model mostly outperforms AFET on the Wiki and BBN datasets. However, it performs substantially worse on the Ontonotes dataset. This is most likely due to the high quality of the handcrafted features present in Ontonotes which help to boost AFET's performance.

\begin{table*}[!t]
\centering
\begin{tabular}{ c | c | c | c | c | c | c | c | c | c | c}

\multirow{2}{*}{\textbf{  Embeddings  }}& \multicolumn{3}{c|}{\textbf{M-Wiki}}  & \multicolumn{3}{c|}{\textbf{M-Ontonotes}}  & \multicolumn{3}{c}{\textbf{M-BBN}} \\ \cline{2-9}
   & \cellcolor{gray!10}{Acc} & \cellcolor{gray!10}{Ma-F1} & Mi-F1 & \cellcolor{gray!10}{Acc} & \cellcolor{gray!10}{Ma-F1} & Mi-F1 & \cellcolor{gray!10}{Acc} & \cellcolor{gray!10}{Ma-F1} & Mi-F1 \\ \hline

Uniform  & \cellcolor{gray!10}{0.850} & \cellcolor{gray!10}{0.017} & 0.188 & \cellcolor{gray!10}{0.632} & \cellcolor{gray!10}{0.265} & 0.511 & \cellcolor{gray!10}{0.905} & \cellcolor{gray!10}{0.054} & 0.526 \\ \hline
GloVe    & \cellcolor{gray!10}{0.899} & \cellcolor{gray!10}{0.028} & 0.358 & \cellcolor{gray!10}{0.708} & \cellcolor{gray!10}{0.381} & 0.683 & \cellcolor{gray!10}{0.943} & \cellcolor{gray!10}{0.085} & 0.754  \\ \hline
Word2Vec & \cellcolor{gray!10}{0.892} & \cellcolor{gray!10}{\textbf{0.032}} & 0.379 & \cellcolor{gray!10}{0.704} & \cellcolor{gray!10}{0.376} & 0.692 & \cellcolor{gray!10}{0.941} & \cellcolor{gray!10}{0.079} & 0.734 \\ \hline
BERT     & \cellcolor{gray!10}{\textbf{0.905}} & \cellcolor{gray!10}{0.029} & \textbf{0.414} & \cellcolor{gray!10}{\textbf{0.787}} & \cellcolor{gray!10}{\textbf{0.412}} & \textbf{0.768} & \cellcolor{gray!10}{\textbf{0.975}} & \cellcolor{gray!10}{\textbf{0.101}} & \textbf{0.893} \\ \hline
%BERT & 0.923 & 0.032 & 0.424 & 0.742 & 0.374 & 0.713 & 0.959 & 0.089 & 0.819 \\ \hline

\end{tabular}
\caption{\label{tab:performanceanalysis-e2e} The results of E2EET model on the modified Wiki, Ontonotes and BBN datasets using various embedding techniques. Misleading metrics are in grey.
  }
\end{table*}

\begin{table*}[!t]
\centering
\begin{tabular}{l | l | c | c | c | c | c | c | c | c | c}

\multirow{2}{*}{\textbf{Model}} & \multirow{2}{*}{\textbf{Attention}} & \multicolumn{3}{c|}{\textbf{M-Wiki}}  & \multicolumn{3}{c|}{\textbf{M-Ontonotes}}  & \multicolumn{3}{c}{\textbf{M-BBN}} \\ \cline{3-11}
  & & Acc & Ma-F1 & Mi-F1 & Acc & Ma-F1 & Mi-F1 & Acc & Ma-F1 & Mi-F1 \\ \hline
%\multicolumn{5}{c}{Word-level model} \\

%AFET                    & -.--- & -.--- & -.--- & -.--- & -.--- & -.--- & -.--- & -.--- & -.---  \\
%NFGET                   & 0.444 & 0.663 & 0.658 & 0.726 & 0.891 & 0.847 & -     & -     & -      \\

\multirow{3}{*}{ML} & None             & 0.448          & 0.674          & 0.658          & 0.637          & 0.843          & 0.788          & 0.688          & 0.827 & 0.836 \\ \cline{2-11}
& Scalar           & \textbf{0.458} & 0.677          & 0.662          & 0.664          & 0.859          & 0.806          & 0.765          & 0.864 & 0.872 \\ \cline{2-11}
&  Dynamic            & 0.451          & \textbf{0.685} & \textbf{0.662} & \textbf{0.672} & \textbf{0.861} & \textbf{0.808} & 0.780 & 0.864 & 0.873\\ \hline

%Ours                    & 0.458 & 0.677 & 0.662 & 0.672 & 0.865 & 0.813 & 0.765 & 0.864 & 0.872  \\
%E2E   & n/a             & 0.336 & 0.336 & 0.431 & 0.610 & 0.729 & 0.749 & 0.702 & 0.776 & 0.844 \\ \hline
E2E   & n/a             & 0.180 & 0.307 & 0.456 & 0.668 & 0.803 & 0.795 & \textbf{0.834} & \textbf{0.888} &  \textbf{0.916} \\ \hline
%Predictable             & 0.903 & 0.148 & 0.452 & 0.742 & 0.396 & 0.713 & 0.960 & 0.429 & 0.822   \\  \hline
%Fil. + Pred.            & 0.318 & 0.154 & 0.517 & 0.610 & 0.406 & 0.749 & 0.705 & 0.444 & 0.847   \\

\end{tabular}
\caption{\label{tab:performanceanalysis-mlvse2e} A comparison between our mention-level (ML) model (with three different attention mechanisms), and E2EET on the modified Wiki, Ontonotes and BBN datasets. Top scores (prior to rounding) are in bold.% Filtering has been applied to the predicted labels and ground truth labels of the E2E model.
  }
\end{table*}

The attention mechanisms (scalar and dynamic) generally had very little impact on performance. Additionally, despite its complexity, the dynamic attention mechanism was often outperformed by the scalar variant. It was found during training that the dynamic attention mechanism performs best on the validation sets (which comprise 10\% of the training data), but there was a significant difference (0.2) in F1 scores between the validation and test sets. The most likely cause of this phenomenon is that the training and testing sets are too distinct from one another, leading to rapid overfitting in more complex models such as attention-based models. This is supported by the fact that the training sets of BBN and Ontonotes were automatically labelled, whereas the testing sets were manually labelled~\cite{shimaoka2016neural}. %

The results clearly show that our mention-level entity typing model outperforms all current state-of-the-art techniques that do not rely on handcrafted features. It is also highly competitive with AFET, a top-performing system that uses handcrafted features.

%There does not appear to be a strong relationship between the attention mechanism used in our model and performance. Each attention mechanism performed well in different settings, with the no-attention variant tending to slightly outperform the others overall. While these results may hint at the conclusion that the attention mechanism has no effect on performance, our investigation in Section~\ref{subsubsec:e2ebaseline}.

\subsection{End-to-end entity typing (E2EET) performance}

\subsubsection{Baseline performance}
\label{subsubsec:e2ebaseline}
To evaluate E2EET we test the model after its embedding layer has been initialised with four different embedding techniques. Table~\ref{tab:performanceanalysis-e2e} shows the results of E2EET on the modified Wiki, Ontonotes and BBN datasets. Here, the strict accuracy, macro, and micro F1 scores are calculated with respect to the model's predictions of every single token in the corpus.

The BERT embeddings significantly outperform the other embedding techniques on every dataset. It is clear that the context-dependent embeddings provided by the BERT model are highly effective when used to support an entity typing model.

The results indicate that the accuracy and macro-F1 metrics used for mention-level models do not accurately reflect the performance of an end-to-end model. The accuracy is extremely high because the vast majority of tokens are not entities, and the model successfully predicts no labels for these tokens. Macro-F1 is similarly misleading, being consistently low due to the scores being divided by the total number of tokens. The only useful metric for evaluating E2EET appears to be micro F1, which disregards non-entities by dividing by the sum of ground truth labels for each token.

Disregarding the misleading accuracy and macro-F1 scores, E2EET performs well on the M-Ontonotes and M-BBN datasets. It did not fare  well on the M-Wiki dataset, however, as it is considerably noisier. Overall, the results indicate that the model is capable of performing end-to-end entity typing, but future research regarding end-to-end models should investigate and devise more suitable evaluation metrics.

\subsubsection{Comparison with mention-level model}
\label{subsubsec:e2evsml}

Table~\ref{tab:performanceanalysis-mlvse2e} shows a comparison between our mention-level and E2EET when trained and evaluated upon the modified datasets. Here, E2EET is using BERT embeddings and is evaluated using the same F1 calculations as the mention-level model, i.e. the scores are only calculated across entities as opposed to across all tokens. This means that any non-entities that are incorrectly labelled as entities by E2EET are ignored.

E2EET is competitive with the mention-level model on the relatively clean M-Ontonotes and M-BBN datasets, even outperforming the mention-level model on M-BBN. This is surprising considering E2EET has a vastly more difficult training objective, and does not know which tokens are entity mentions. The most likely explanation for this result is that the context of the entire sequence plays a pivotal role in the model's success, particularly in M-BBN. It allows E2EET to more effectively classify each token than the mention-level model which was trained on smaller context windows.

Another noticeable result is that, for the mention-level typing models, there is a clear relationship between the complexity of the attention model and the overall performance. This is in contrast to the results on the original datasets (Table~\ref{tab:performancecomparison}), where the attention mechanism had little impact on performance. The dynamic attention model in particular excels on the modified datasets, which are significantly cleaner than the original datasets as a result of being taken from the hand-labelled test set of their respective original dataset. \textit{The dynamic attention mechanism is clearly most effective when the dataset is clean and when there is consistency between the training and testing sets. }

Overall the results of the comparison shows that E2EET is highly competitive with the mention-level model as a result of its ability to incorporate document-level context. It performs well on clean datasets where there is little disparity between the training and evaluation sets, and provides a strong foundation for future research into entity typing.

\section{Conclusion}

In this paper we have carried out an extensive ablative study demonstrating the effectiveness of contextualised embeddings for mention-level entity typing and have shown the competitiveness of our proposed end-to-end system for entity typing. Our \textit{mention-level} model embeds the left, right, and mention contexts using BERT and employs two novel attention mechanisms in order to predict the labels associated with each entity mention. Our \textit{end-to-end} model (E2EET), on the other hand, effectively determines the type(s) of all tokens in a sentence. Results show that our mention-level model outperforms state-of-the-art mention-level entity typing models. Our end-to-end model performs well on clean datasets and is capable of outperforming the mention-level model despite not knowing which tokens in each sentence are entities.

In future we plan to run ten-fold cross validation to ensure statistical significance in the experiments. It would be interesting to investigate the effectiveness of other recent context-dependent embeddings such as XLNet~\cite{yang2019xlnet}, transformers to replace Bi-GRU for context representation learning, and to incorporate hierarchical encoding techniques to improve prediction accuracy.

%\begin{acks}
%This research was funded by an Australian Postgraduate Award Scholarship and a UWA Safety Net Top-up Scholarship.
%\end{acks}

\bibliography{main}
\bibliographystyle{splncs04} 

\end{document}